\documentclass[letterpaper, 10 pt, conference]{ieeeconf}

\IEEEoverridecommandlockouts
\usepackage{graphicx}       % Preferred package to inclued figures.
\usepackage{booktabs}       % Preferred package to create tables.
\usepackage{amsmath}        % Extends the available math styles and symbols.
\usepackage{microtype}      % For nerds.
\usepackage[utf8]{inputenc} % Directly enter non-ascii characters.
\usepackage{newtxmath}      % Times font in math mode. Be consistent with the typical times text font.
\usepackage[acronym,shortcuts]{glossaries} % Declaring acronyms at first use.
\usepackage{hyperref}       % URLs

\renewcommand{\vec}[1]{\boldsymbol{#1}}   % Command for vectors.
\newcommand{\mat}[1]{\boldsymbol{#1}}     % Command for matrices.
\newcommand{\T}{^{\mathsf{T}}}            % Command for transpose.
\newcommand{\unit}[1]{\,\mathrm{#1}}      % Command for units (upright, small space before unit).

\newcommand{\Equation}[1]{(\ref{#1})}
\newcommand{\Figure}[1]{Fig. \ref{#1}}

\newcommand{\Section}[1]{Sec. \ref{#1}}

\newacronym{asl}{ASL}{Autonomous Systems Lab}
\newacronym{asctec}{AscTec}{Ascending Technologies}
\newacronym{cpp}{CPP}{coverage path planning}
\newacronym{dof}{DoF}{degree of freedom}
\newacronym{dz}{DZ}{dropping zone}
\newacronym{ekf}{EKF}{extended Kalman filter}
\newacronym{enu}{ENU}{east north up}
\newacronym{epm}{EPM}{electropermanent magnet}
\newacronym{eurathlon}{euRathlon}{European Robotics League}
\newacronym{euroc}{EuRoC}{European Robotics Challenges}
\newacronym{fsm}{FSM}{finite state machine}
\newacronym{imu}{IMU}{inertial measurement unit}
\newacronym{ipp}{IPP}{informative path planning}
\newacronym{fov}{FOV}{field of view}
\newacronym{lz}{LZ}{landing zone}
\newacronym{kf}{KF}{Kalman filter}
\newacronym{mav}{MAV}{micro aerial vehicle}
\newacronym{mbzirc}{MBZIRC}{Mohamed Bin Zayed International Robotics Challenge 2017}
\newacronym{mcu}{MCU}{microcontroller unit}
\newacronym{nmpc}{NMPC}{nonlinear model predictive position control}
\newacronym{msf}{MSF}{Multi Sensor Fusion}
\newacronym{pcb}{PCB}{printed circuit board}
\newacronym{pwm}{PWM}{pulse-width modulation}
\newacronym{rmse}{RMSE}{root-mean-square error}
\newacronym{sar}{SAR}{search and rescue}
\newacronym{sdk}{SDK}{software development kit}
\newacronym{vio}{VIO}{visual-inertial odometry}
\newacronym{vs}{VS}{visual servoing}

\title{\LARGE\bf A Decentralized Multi-Agent Unmanned Aerial System to \\
Search, Pick Up, and Relocate Objects}

\author{Rik Bähnemann, Dominik Schindler, Mina Kamel, Roland Siegwart, and Juan Nieto
\thanks{R. Bähnemann, M. Kamel, R. Siegwart, and J. Nieto are with the \ac{asl}, ETH Zürich, Zürich, Switzerland.
D.~Schindler was also with the \ac{asl} at the time this work was conducted.
\newline \tt \{brik, schidomi, rsiegwart, nietoj\}@ethz.ch}}

\begin{document}
\maketitle
\thispagestyle{empty}
\pagestyle{empty}
\begin{abstract}
We present a fully integrated autonomous multi-robot aerial system for finding and collecting moving and static objects with unknown locations.
This task addresses multiple relevant problems in \ac{sar} robotics such as multi-agent aerial exploration, object detection and tracking, and aerial gripping.
Usually, the community tackles these problems individually but the integration into a working system generates extra complexity which is rarely addressed.
We show that this task can be solved reliably using only simple components.
Our decentralized system uses accurate global state estimation, reactive collision avoidance, and sweep planning for multi-agent exploration.
Objects are detected, tracked, and picked up using blob detection, inverse 3D-projection, Kalman filtering, visual-servoing, and a magnetic gripper.
We evaluate the individual components of our system on the real platform.
The full system has been deployed successfully in various public demonstrations, field tests, and the \ac{mbzirc}.
Among the contestants we showed reliable performances and reached second place out of 17 in the individual challenge.
\end{abstract}

\section{Introduction}
Rotary-wing \acp{mav} are becoming increasingly popular in disaster response scenarios such as floods, earth-quakes, or wild fires.
Due to their growing availability and excellent camera guidance capabilities \acp{mav} are frequently employed to extend the human eye in damage assessment, area mapping, and visual inspection \cite{qi2016search,de2014uas,kruijff2016deployment}.
Although researchers have put great effort to extend the range of \ac{mav} capabilities, e.g., autonomous exploration, aerial gripping, or transportation \cite{maza2007multiple, lindsey2011construction, bernard2011autonomous}, their application is still limited to human operated inspection and filming.
We wish to extend the application field by showing that existing technologies for aerial gripping can be merged into a deployable autonomous system.

\begin{figure}
  \centering
  \includegraphics[width=0.48\textwidth]{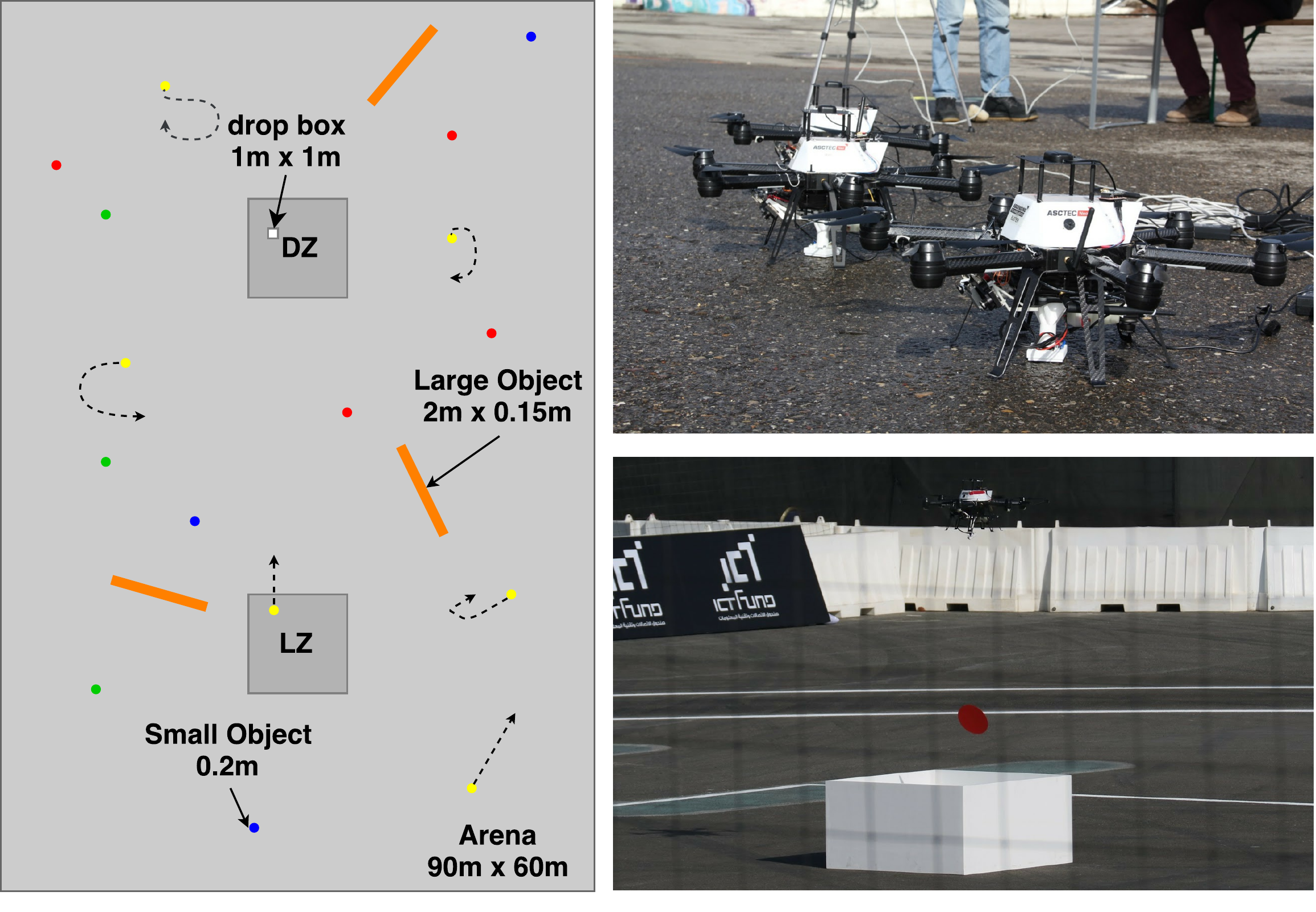}
  \caption{Example scenario \ac{mbzirc}: Starting from the \acf{lz} our three \acp{mav} need to collect all objects in the arena and deliver them to the \acf{dz} or drop box.
  The top right image shows our platforms consisting of three identical \ac{asctec} Neo hexacopters equipped a custom sensor suite.
  The bottom right image shows a successful delivery during a trial.}
  \label{fig:arena}
\end{figure}

In this work we present a fully autonomous multi-agent aerial system to search and pick up small objects in environments similar as depicted in \Figure{fig:arena} which shows the complete system deployed at \ac{mbzirc}.
The system consists of multiple building blocks relevant to \ac{sar} scenarios.
The first block is \emph{multi-agent aerial exploration} which includes agent allocation, coverage planning, globally consistent state estimation, and collision avoidance.
The second block is \emph{aerial gripping} which contains object detection, tracking, visual servoing, and physical interaction.

\emph{Multi-agent aerial exploration} in an outdoor environment has four key requirements.
It needs to cover the full area, be collision free, use limited network bandwidth, and be robust to agent failure.
Thus we propose a decentralized system in which each agent is capable of fulfilling the task independently and share a minimum of information.
Each agent is assigned to a predefined area following a camera coverage pattern.
\Ac{vio} fused with RTK-GPS gives accurate, high bandwidth, fail-safe state estimation in a common coordinate frame.
The \acp{mav} broadcast their position and velocity over the network which allows other agents to avoid them.
This reactive avoidance scheme enables each agent to move freely without knowing the intention of the others explicitly.

The \emph{aerial gripping} pipeline is based on object servoing which is independent of the underlying state estimation.
Visual servoing allows correct positioning of the \ac{mav} relative to a target object without external information about its position.
In particular, we present an object detection pipeline to retrieve the 3D position of objects from a single monocular detection and a simple yet precise servoing algorithm.
The aerial gripping is complemented by an energy-saving, compliant \ac{epm} gripper design.
Unlike regular electromagnets, \acp{epm} only draw electric current while transitioning between the states.

\begin{figure}
  \centering
  \includegraphics[width=0.48\textwidth]{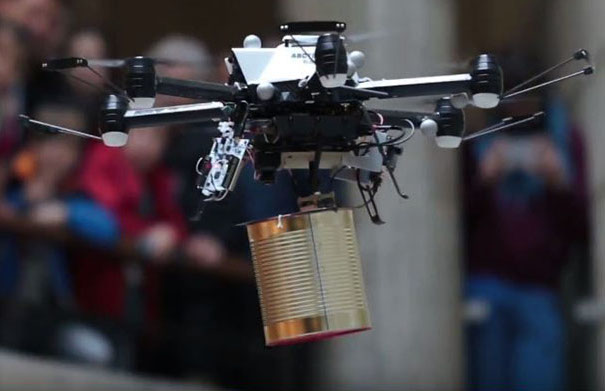}
  \caption{Public demonstration of our transportation system\setcounter{footnote}{0}\protect\footnotemark.
  The \ac{mav} autonomously explores, finds, and picks up objects from locations inaccessible from the ground. (Photo: ETH Zürich)}
  \label{fig:eyecatcher}
\end{figure}
\footnotetext{Video of public demonstration: \url{https://youtu.be/sk0XZ01Paqw}}

The main contributions are
\begin{itemize}
  \item a modular, decentralized, collision-free multi-agent aerial search, pick up and delivery system,
  \item formulation and evaluation of a complete image to position commands servoing pipeline for static and moving objects, and
  \item evaluation and deployment of the system in different environments.
\end{itemize}

We organize the paper as follows.
Section \ref{sec:related_work} presents related work.
In section \ref{sec:approach} we summarize our multi-agent system architecture and exploration strategy.
In section \ref{sec:servoing} we present the aerial gripping pipeline.
The system hardware is described in the implementation details \ref{sec:implementation}.
Finally, we validate our methods in section \ref{sec:results} before we close the article with concluding remarks.

\begin{figure*}
  \centering
  \includegraphics[width=\textwidth]{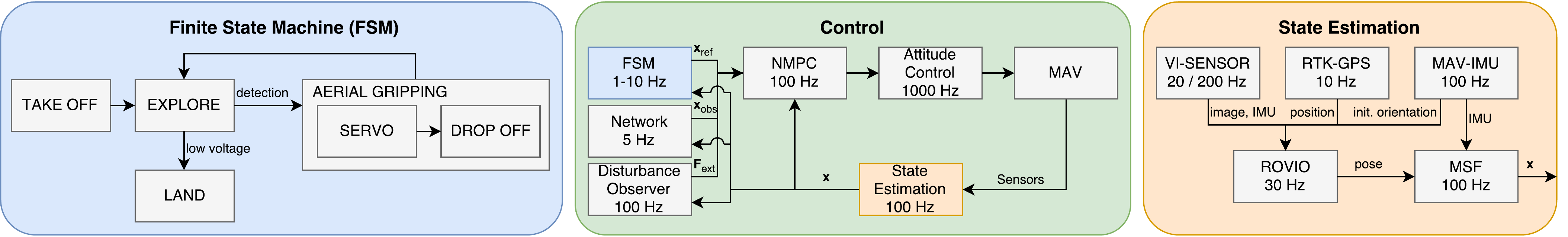}
  \caption{System overview: the \ac{fsm} of a decentralized agent (blue), the control pipeline (green), and the state estimation pipeline (orange).
  The \ac{fsm} sends reference trajectories $\vec{x}_{\mathrm{ref}}$ to the position controller.
  The position controller uses the own global \ac{mav} state $\vec{x}$, the state of other agents $\vec{x}_\mathrm{obs}$, and estimated disturbance forces $\vec{F}_\mathrm{ext}$ to provide collision- and offset-free trajectory tracking.
  The state estimation fuses \ac{vio} with RTK-GPS to retrieve a global state $x$.}
  \label{fig:system}
\end{figure*}

\section{Related Work}
\label{sec:related_work}
This work is a continuation of our previous work on aerial pick-up and delivery of magnetic objects \cite{Gawel16}.
While the previous work focused on developing a first running single prototype, this work presents a fully integrated multi-agent system with an improved gripper design and a new formulation of the servoing problem.
In \cite{Gawel16} we calculated the object position based on the \ac{mav} state relative to the object, here we calculate it solely from the detection.

Our proposed system follows similar design principles as other successful teams in \ac{mbzirc}.
As in \cite{nieuwenhuisen2017} we pick components that are easily adaptable, robust, and whose behavior can be interpreted and tuned by the operator.

We focus the rest of the review on the two main areas of our system: \emph{multi-robot aerial exploration} and \emph{aerial gripping}.
\subsection{Multi-Robot Aerial Exploration}
Multi-robot 2D exploration in known environments is a common problem in outdoor surveying.
Geometric solutions are complete, deterministic, computationally fast, and easily tractable.
\cite{barrientos2011aerial} discretize an area into viewpoints, assign each \ac{mav} to a subset of regions and use a wavefront planner for coverage.
\cite{maza2007multiple} use polygonal area decomposition and \ac{fov} aware sweep planning to plan paths for cooperative \ac{mav} search operations.
We implement a specific case of \cite{maza2007multiple}'s \ac{cpp}, where we manually divide the workspace and have a fixed \ac{fov}.

In contrast to existing multi-robot systems, which usually assume collision-free planning by construction, we focus on the practical implications the decentralized picking scenario in a heterogeneous robot environment introduces: accurate global positioning and reactive collision avoidance in case of agent interference.
For global positioning we fuse ROVIO \cite{bloesch2015robust} with RTK-GPS which gives more accurate longitudinal and lateral accuracy, reliable altitude information and fall back \ac{vio} positioning compared to standard IMU-GPS solutions \cite{wendel2006integrated}.
For collision avoidance we choose reactive avoidance at the control level \cite{kamel2017nonlinear}.
In comparison to centralized global planners like \cite{augugliaro2012generation}, this reactive approach requires significantly less communication and is safely embedded in the control framework.

\subsection{Aerial Gripping}
Aerial gripping is a quickly growing field in \ac{mav} research.
Different approaches exist both for the gripping mechanism as well as servoing and transportation.
In \cite{ghadiok2011autonomous} the authors present an integrated object detection and gripping system.
They detect static objects using infrared LEDs and grab them using a form-closure mechanical design.
In contrast, our system performs visual servoing from RGB images on both static and moving objects and aims at gripping ferrous objects with a magnetic gripper.
In \cite{thomas2016visual} the authors use a claw-like gripper to bring a \ac{mav} to a perching position, hanging from a pole.
They use image based visual servoing.
The control law is based entirely on the error in the image plane, no object pose estimation is performed.
In contrast, we use pose based visual servoing.
In this approach, the object pose is estimated from the image stream, then the robot is commanded to move towards the object to perform grasping.
In \cite{lindsey2011construction,augugliaro2014flight} the authors present multi-robot aerial systems for autonomous structure assembly.
Both works have knowledge of object positions and operate open loop and centralized in a Vicon environment.
We perform visual servoing to grasp objects under outdoor conditions in a decentralized system.
%Different gripping mechanisms like suction, penetration, robotic arms or claws are presented in \cite{kessens2016versatile,mellinger2011design,kim2013aerial,thomas2013avian}.
%We use a passively compliant \ac{epm}.
%Furthermore, we focus our work on detection and servoing in outdoor environments.

\section{Multi-robot Aerial Exploration}
\label{sec:approach}
Our system architecture is visualized in \Figure{fig:system}.
The robot's brain is a decentralized \acf{fsm} that handles higher-level, single-agent task allocation and generates reference trajectory commands.
The control loop processes trajectory commands and handles external disturbances and reactive collision avoidance.
The state estimation pipeline provides the robot and other agents with consistent, global states.
Each block in the system has a defined input and output and can be refined into smaller units, modularly replaced by advanced solutions, and tested separately.
In the following we detail the three main blocks.

\subsection{Decentralized Finite State Machine (Blue)}
Each agent acts independently of the other agents and all algorithms run onboard the \ac{mav}.
After take-off it alternates between exploring a predefined area and greedily picking up and delivering objects.
The robot lands when its battery voltage is below some threshold and it is currently not attempting to deliver an object.

\begin{figure}
  \centering
  \includegraphics[width=0.48\textwidth]{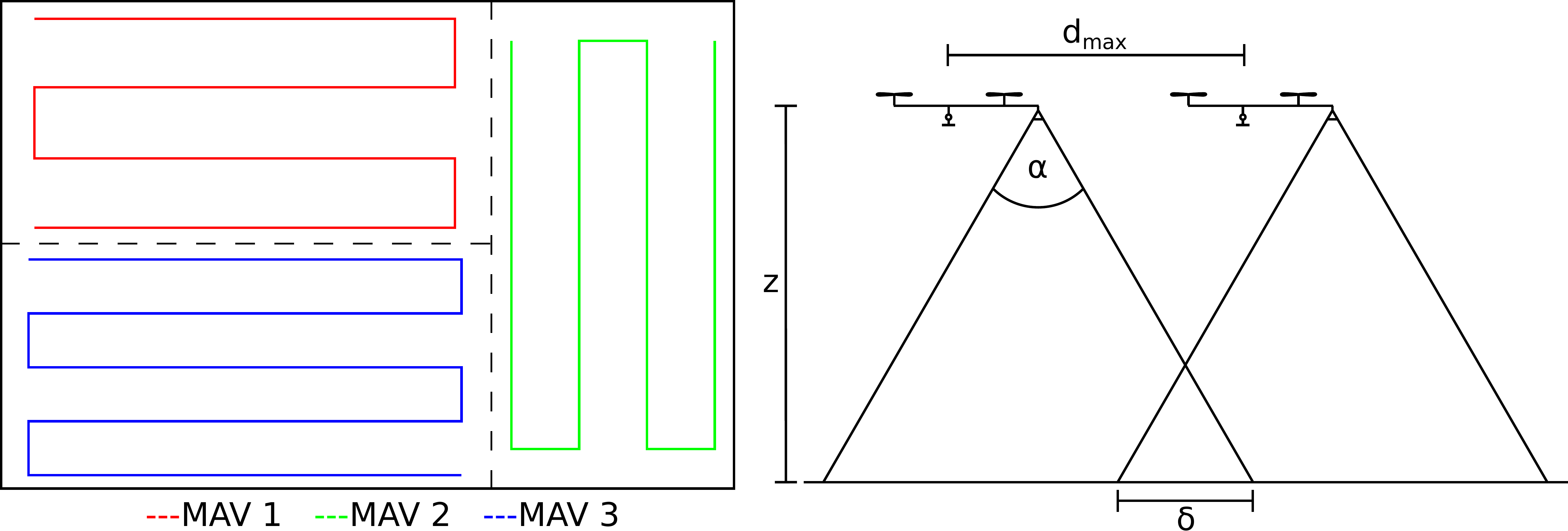}
  \caption{\ac{cpp}: The search region is manually divided into convex regions.
  Each \ac{mav} explores its region with a zig-zag path.
  The maximum sweep distance $d_{max}$ is a function of altitude $z$, camera \ac{fov} $\alpha$, and overlap $\delta$.}
  \label{fig:exploration}
\end{figure}

To explore the arena, we separate it into three convex regions and implement a sweep planner as depicted in \Figure{fig:exploration}.
The maximum distance $d_{max}$ between two line sweeps is calculated based on the camera's \ac{fov} $\alpha$, the \ac{mav} altitude $z$, and a user defined view overlap $\delta \in [0\ldots1]$.
\begin{align}
  d_{max} = (1 - \delta) \cdot 2 \cdot z \cdot \tan{\frac{\alpha}{2}}
\end{align}

The field division creates a first safety layer for collision avoidance.
However, interference between the agents can still occur in pick up situations or with other agents in a heterogeneous system.
Therefore, we provide a second safety layer at the control level.

\subsection{Reactive and Adaptive Control (Green)}
The control pipeline is based on a standard cascaded control scheme where a slower outer \ac{nmpc} loop controls a fast attitude control loop \cite{achtelik2011onboard}.
The position controller has three functionalities.
It tracks the reference trajectory from the \ac{fsm} requesting position and velocity commands, it compensates for changes in mass and wind with an \ac{ekf} disturbance observer, and it uses other agents states for reactive collision avoidance \cite{kamel2016linear,kamel2017nonlinear}.
For the latter, the controller includes obstacles as a hard and soft constraint into the trajectory tracking optimization \cite{kamel2017nonlinear}.
This paradigm can include arbitrary agent's states and always ensure that a minimum distance is kept.
Thus also other robots or obstacles can be avoided as long as they broadcast their state in a common frame over WiFi.
Essentially, this is the only communication on the network.

\subsection{Global State Estimation (Orange)}
In order to operate multiple agents in the same area they need to share positions in a common global coordinate frame.
For this we integrate RTK-GPS position updates into our \ac{vio} pose estimation framework ROVIO \cite{bloesch2015robust}.
This gives us drift-free, high-bandwidth, fail-safe, accurate global positioning.

The signal flow is depicted in the orange frame in \Figure{fig:system}.
In a first step we initialize the \ac{vio}'s pose to the current GPS position and magnetometer orientation.
During operation the framework then internally uses the position updates to correct the its pose estimate from \ac{vio}.
Finally, \ac{msf} fuses the global pose from ROVIO with the onboard \ac{imu} \cite{lynen13robust}.

\section{Aerial Gripping}
\label{sec:servoing}
A key component of our system is the object tracking and gripping pipeline.
The \acp{mav} need to accurately detect and servo to small moving and stationary discs from $5$ to $10 \unit{m}$ altitude.

\subsection{Object Tracking Pipeline}
\begin{figure}
  \centering
  \includegraphics[width=0.48\textwidth]{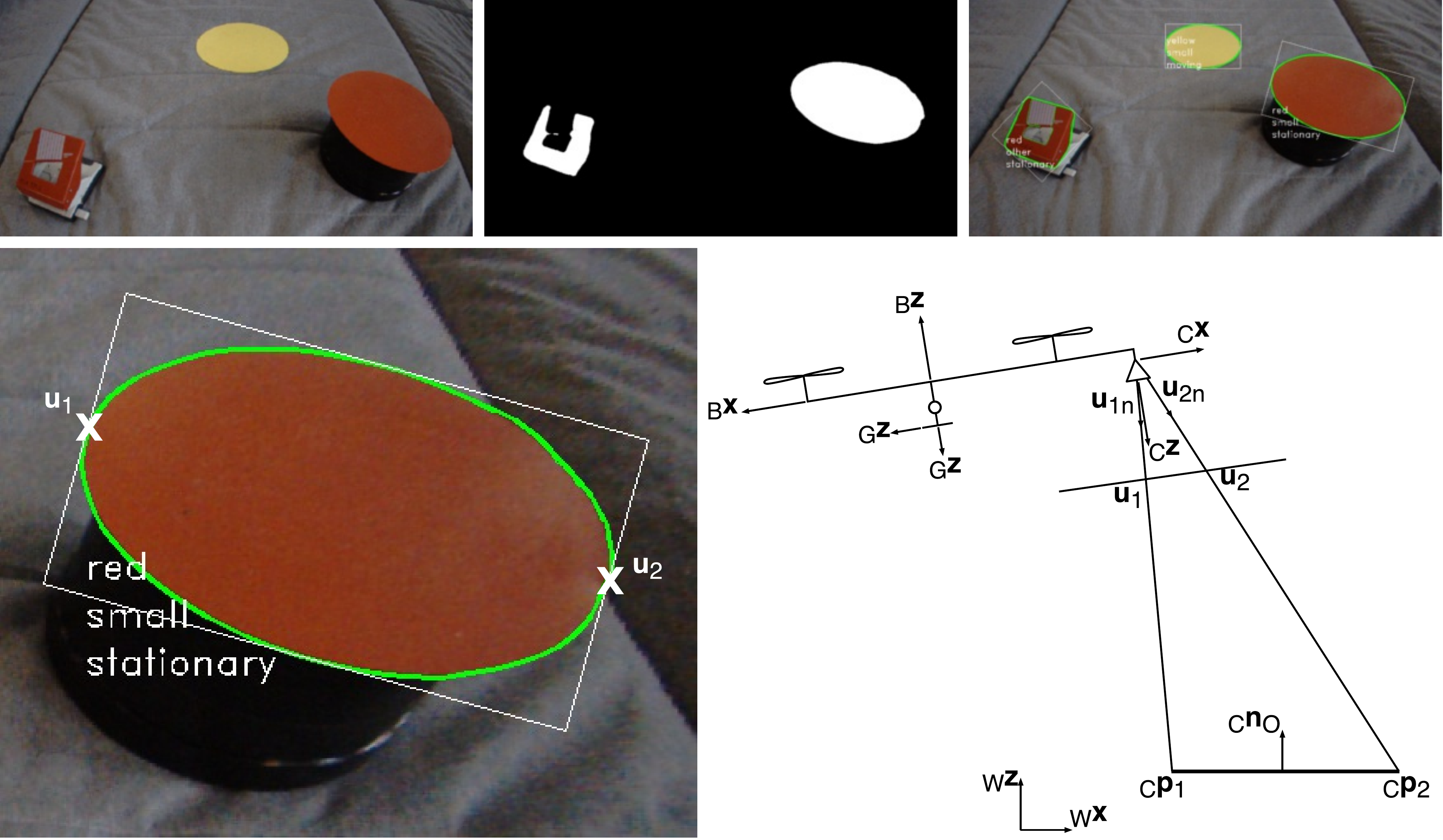}
  \caption{The object tracking pipeline: (top) input image, binary image for red color, detections with classifications.
  (bottom) The 3D positions of the object corners in camera frame $_C\vec{p}_1$ and $_C\vec{p}_2$ can be calculated from the projected points $\vec{u}_1$ and $\vec{u}_2$ in the image plane given the size and planar positioning of the object.}
  \label{fig:image_processing}
\end{figure}
In order to locate colored objects, time-stamped RGB images are fed into our object detection and tracking pipeline.
The pipeline outputs the 3D positions and the 2D horizontal velocities of the observed objects.
The detection is visualized in \Figure{fig:image_processing} and the main steps of the pipeline are:

\begin{enumerate}
\item Undistort the image and convert its pixel values from RGB to \emph{CIE L*a*b*} color space for intuitive thresholding.
\item Get the \emph{detections} as the thresholded image regions.
\item Classify the shapes based on \emph{contour properties}.
\item Remove \emph{outlier detections} by their classification, shape-color inconsistency, and flight altitude-size inconsistency.
\item Compute the \emph{inverse projection} of the detections from 2D to 3D in camera coordinate frame (\Section{sec:inverse-projection}). \label{enum:inverse-projection}
\item Transform the object position from the camera coordinate frame $C$ to the world coordinate frame $W$ using the \ac{mav} pose and extrinsic camera calibration \cite{furgale2013unified}.
\item Initialize a \emph{\ac{kf}} to track the position and velocity.
Assign detections to already initialized \acp{kf} using the \emph{Hungarian algorithm} \cite{kuhn1955hungarian}.
\end{enumerate}

\subsection{Calculate 3D position from a Single Object Detection}
\label{sec:inverse-projection}
\Figure{fig:image_processing} (bottom) displays the problem of calculating the position of two points on an object w.r.t the camera.
To compute the 3D points $_C\vec{p}_1$ and $_C\vec{p}_2$ in the camera coordinate frame $C$, we make two assumptions:
\begin{itemize}
  \item the points lie on a plane perpendicular to the world frame z-axis ${_W\vec{z}}$,
  \item the metric 3D distance $l$ between the two points is known.
\end{itemize}
These assumptions impose two constraints:
\begin{align}
 ||{_C\vec{p}_1} - {_C\vec{p}_2} || &= l, \label{eq:distance} \\
 {_C\vec{n}_O}\cdot\left( {_C\vec{p}_1} - {_C\vec{p}_2} \right) &= 0, \label{eq:perpendicular}
\end{align}
where ${_C\vec{n}_O} = \mat{R}_{CW}\,{_W\vec{z}}$ is the object normal expressed in the camera coordinate frame and $\mat{R}_{CW}$ is the rotation matrix from the camera coordinate frame to the world coordinate frame.
Given the mapped points $\vec{u}_1$ and $\vec{u}_2$ in the image correspond to the points ${_C\vec{p}_1}$ and ${_C\vec{p}_2}$, we write them as
\begin{align}
 {_C\vec{p}_1} &= {\lambda_1} \, l \, \vec{u}_{1n}, \label{eq:lambda_1} \\
 {_C\vec{p}_2} &= {\lambda_2} \, l \, \vec{u}_{2n}, \label{eq:lambda_2}
\end{align}
where $\vec{u}_{1n}$ and $\vec{u}_{2n}$ are arbitrarily scaled vectors pointing from the \emph{focal point} to the points ${_C\vec{p}_1}$ and ${_C\vec{p}_2}$ computed through the \emph{perspective projection} of the camera.
For a pinhole-model the perspective projection from image coordinates $\begin{pmatrix}u_{x}&u_{y}\end{pmatrix}\T$ to image vector $\begin{pmatrix}u_{nx}&u_{ny}\end{pmatrix}\T$ is
\begin{align}
  u_{nx} &= \frac{1}{f_x} ( u_{x} - p_x ), & u_{ny} &= \frac{1}{f_y} ( u_{y} - p_y ),
\end{align}
where $p_x$, $p_y$ is the principal point and $f_x$, $f_y$ is the focal length obtained from the intrinsic calibration \cite{furgale2013unified}.

${\lambda_1}$ and ${\lambda_2}$ are two unknown scaling factors.
Inserting \Equation{eq:lambda_1} and \Equation{eq:lambda_2} into \Equation{eq:distance} and \Equation{eq:perpendicular} and solving for ${\lambda_1}$ and ${\lambda_2}$  yields the scaling factors which allow the inverse projection from 2D to 3D:
\begin{align}
  \lambda_1 = \frac{\lvert{_C\vec{n}_O} \cdot {\vec{u}_{2n}}\rvert}{
  \lVert\left({_C\vec{n}_O} \cdot {\vec{u}_{2n}}\right) \, {\vec{u}_{1n}} -
  \left({_C\vec{n}_O} \cdot {\vec{u}_{1n}}\right) \, {\vec{u}_{2n}}\rVert}, \\
  \lambda_2 = \frac{\lvert{_C\vec{n}_O} \cdot {\vec{u}_{1n}}\rvert}{
  \lVert\left({_C\vec{n}_O} \cdot {\vec{u}_{2n}}\right) \, {\vec{u}_{1n}} -
  \left({_C\vec{n}_O} \cdot {\vec{u}_{1n}}\right) \, {\vec{u}_{2n}}\rVert}.
\end{align}
We consider the mean of $_C\vec{p}_1$ and ${_C\vec{p}_2}$ as the object center point in 3D.

\subsection{Object Servoing Pipeline}
\begin{figure*}
  \centering
  \includegraphics[width=\textwidth]{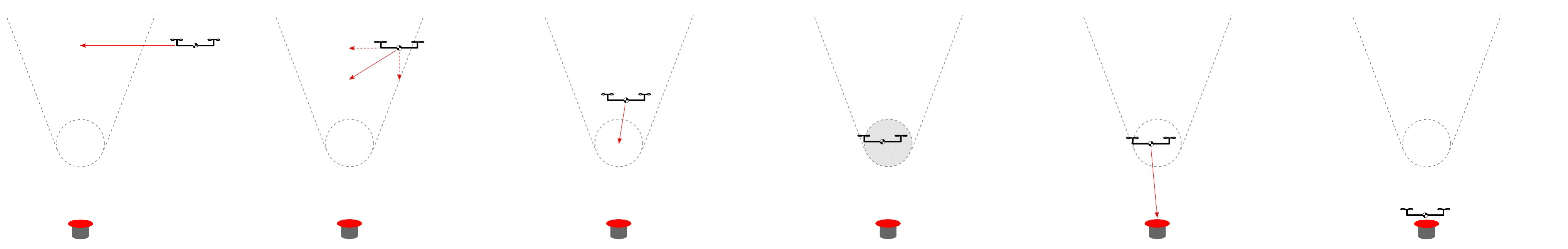}
  \caption{The stages of servoing: descending in a cone, centering and magnet engagement above the object, approach and grasping.}
  \label{fig:servoing}
\end{figure*}
Once the agent detects an object in the exploration phase, it switches into servoing as depicted in \Figure{fig:servoing}.
The servoing algorithm directly sends tracked x-y-object positions and velocities relative to the gripper frame $G$ to the controller.
In order to keep the object in the \ac{fov} and to limit the descending motions, the z-position is constrained such that the \ac{mav} stays in a cone above the object.
When the \ac{mav} is centered in a ball above the object, it activates the magnet and approaches the object using the current track as target position.
During the descent it constantly remagnetizes the gripper to amplify the magnetic force on the object.
The agent senses successful grasping using Hall sensors in the gripper.

\section{Implementation Details}
\label{sec:implementation}
\begin{figure}
  \centering
  \includegraphics[width=0.48\textwidth]{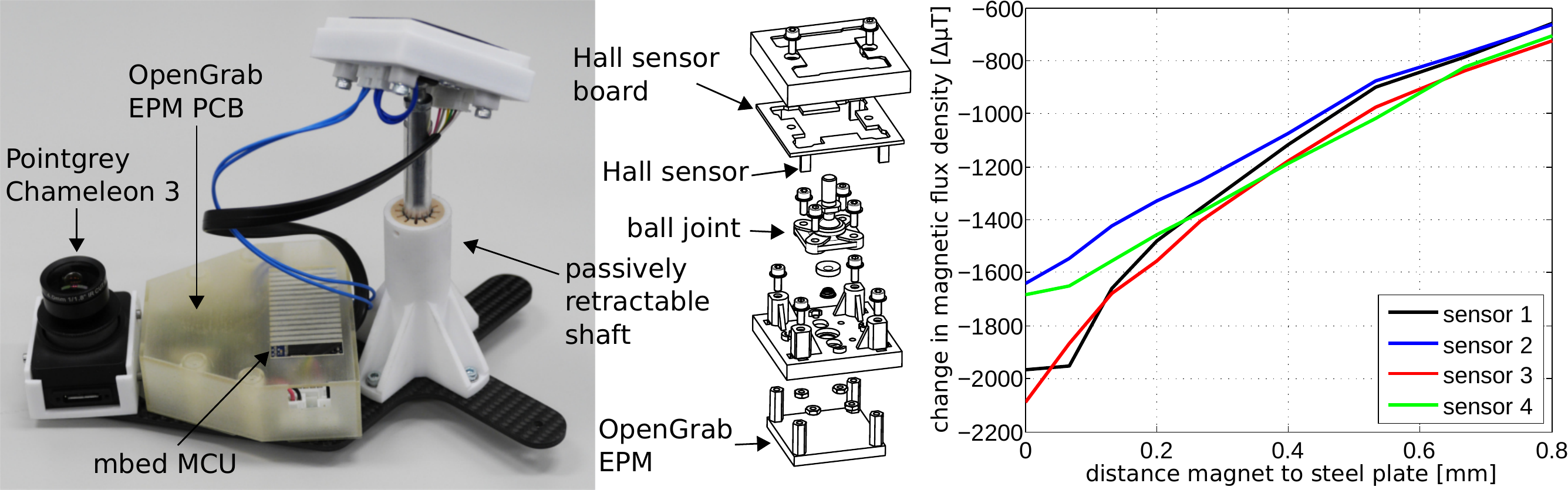}
  \caption{The gripper-camera combination used to detect and pick-up objects. The change in magnetic flux measured by the Hall sensors indicates contact.}
  \label{fig:gripper}
\end{figure}

We run the system on the \ac{asctec} Neo hexacopters shown in \Figure{fig:arena}.
The Neos are equiped with a downward-facing PointGrey/FLIR Chameleon3 color camera ($0.75\unit{MP}$ in experiments, $3.2\unit{MP}$ during \ac{mbzirc}) for object detection, a Skybotix VI-sensor and Piksi RTK-GPS receiver for state estimation, an Intel i5 for onboard processing, and a custom \ac{epm} gripper.
The gripper-camera combination is depicted in \Figure{fig:gripper}.
The gripper is designed to be lightweight, durable, simple, and energy efficient.
Its core module is a NicaDrone \ac{epm} with a typical maximum holding force of $150\unit{N}$ on plain ferrous surfaces.
The \ac{epm} is mounted compliantly on a ball joint on a passively retractable shaft.
The gripper has four Hall sensors placed around the magnet to indicate contact with ferrous objects.
The change in magnetic flux density indicates contact with a ferrous object as shown in \Figure{fig:gripper}.
The total weight of the setup is $250 \unit{g}$.
Unlike our previous version \cite{Gawel16}, the gripper is not compliant to non-planar surface geometries but it has simpler mechanics and its parts are more easily available.

\section{Results}
\label{sec:results}
The partial and full system has been employed in various occasions among which were public demonstrations, field tests, and the \ac{mbzirc} (see \Figure{fig:eyecatcher} and \Figure{fig:arena}).
The modularity of the system allows reformulating the system for different applications.
For example we employed it indoors using vision without GPS, or demonstrated only the aerial gripping.
In the following we evaluate the individual components of the system and report briefly on our \ac{mbzirc} experience.
\subsection{Multi-Robot Aerial Exploration}
In this section we validate the \emph{reactive collision avoidance} and the \emph{state estimation} pipeline.

\begin{figure}
  \centering
  \includegraphics[width=0.48\textwidth]{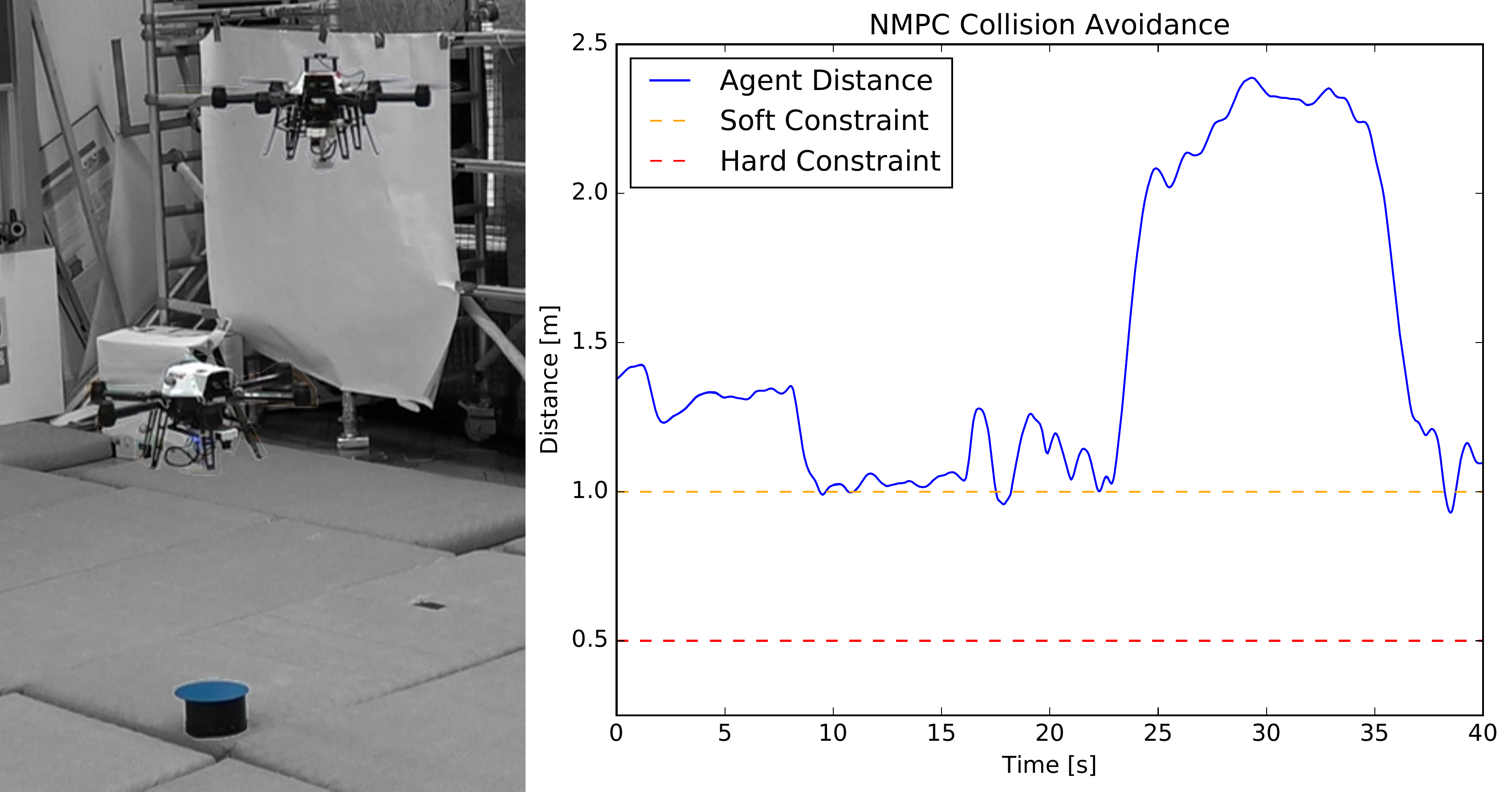}
  \caption{Two drones attempting to servo the same object.
  Sharing the global position over WiFi allows the \ac{nmpc} to prevent collisions without explicitly communicating the agent's intention or desired trajectory.}
  \label{fig:collision}
\end{figure}

To evaluate the \emph{collision avoidance} we let two drones attempt to servo to the same object.
In \Figure{fig:collision} one can observe the distance between the \acp{mav}.
The prioritized \ac{mav} remains in the servoing position.
The less prioritized \ac{mav} tries to enter the space but its position controller keeps it at a predefined distance of $1\unit{m}$.
This allows running different robotic systems in parallel without communicating each others intentions explicitly.

\begin{figure}
  \centering
  \includegraphics[width=0.48\textwidth]{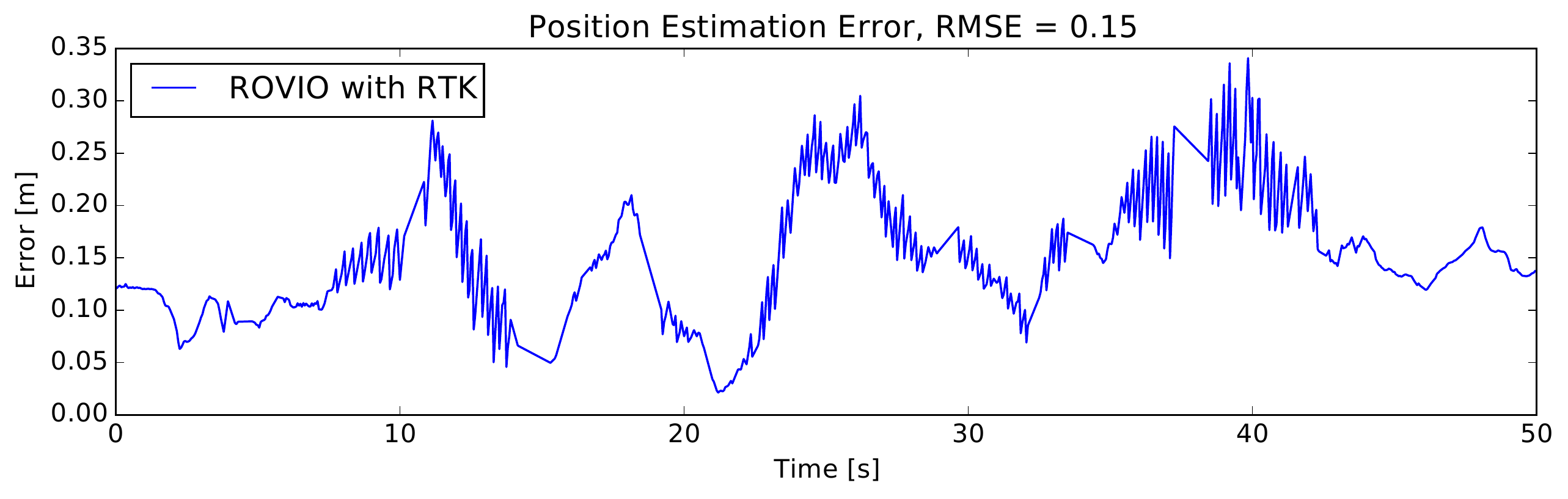}
  \caption{The error between a Leica and a RTK-GPS with ROVIO trajectory.
  The state estimation is accurate enough to use reactive collision avoidance.
  The frequent jumps in the plot are caused by synchronization offset between the groundtruth measurements and state estimation.}
  \label{fig:state_estimation}
\end{figure}

To evaluate the \emph{state estimation} we compare a $90\unit{m}$ trajectory of our \ac{mav} with the ground truth position recorded with a Leica Totalstation in \Figure{fig:state_estimation}.
Since the two trajectories have different time stamps and origins, we manually align them with respect to the \ac{rmse}.
The \ac{rmse} compared to Leica is about $15\unit{cm}$ which allows using the reactive collision avoidance paradigm.


\subsection{Aerial Gripping}
In this section we evaluate the \emph{3D detection accuracy}, the \emph{tracking}, and the \emph{servoing sequence}.
The experiments are conducted in a Vicon motion capture environment for ground truth comparisons.
Note that servoing is independent of the \ac{mav} state estimation.

\begin{figure}
  \centering
  \includegraphics[width=0.48\textwidth]{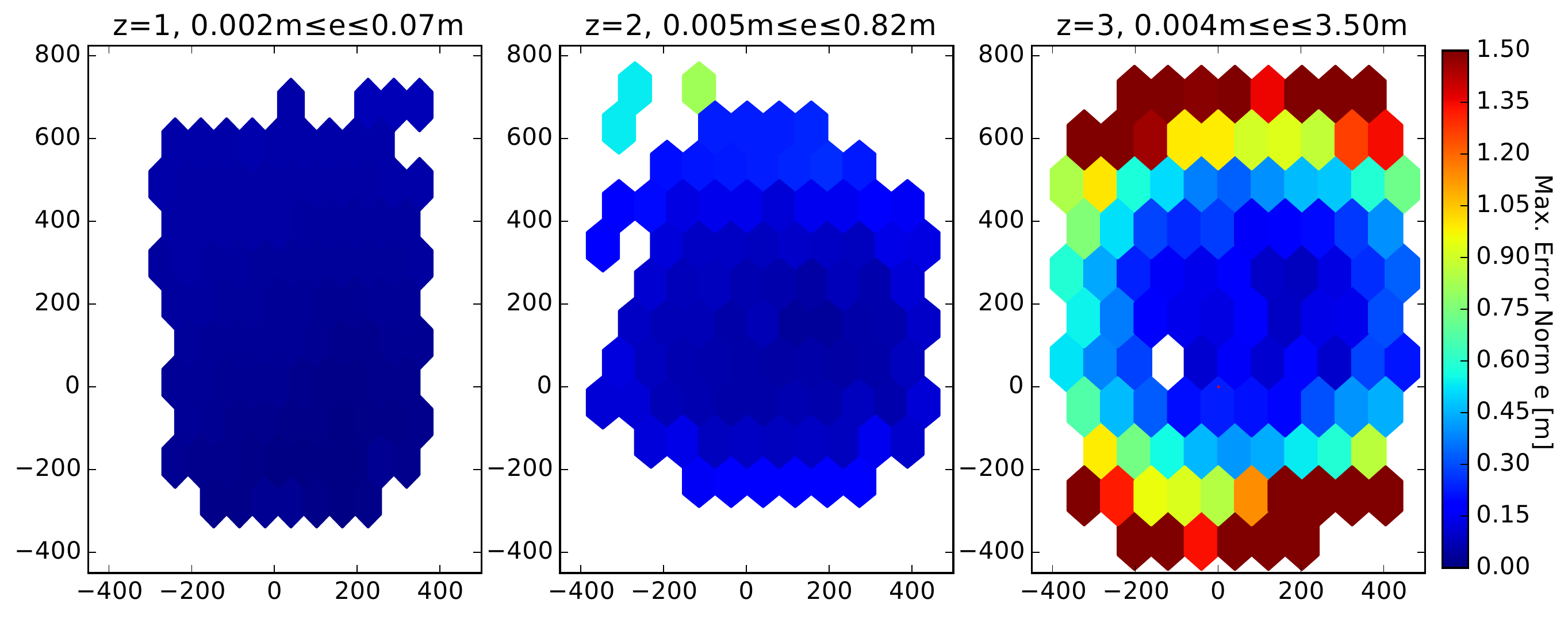}
  \caption{Ground truth validation of the detection error at different altitudes.
  Blank spots mark no detections.
  Centered detections have a position error lower than the measuring accuracy.
  Blob detections at the boundaries are subject to boundary effects like vignetting and have worse accuracy.}
  \label{fig:detection_error}
\end{figure}

\Figure{fig:detection_error} plots the detection error against different object positions in the image plane at different altitudes.
Despite calibration errors and delay, we can achieve a position error of up to $2 \unit{mm}$ which is lower than the measuring accuracy for this setup.
Low resolution and boundary effects like vignetting distort the blob detection and thus lead to larger errors and missing detections (depicted blank) at the image boundary.
This emphasizes the importance of the centering step in the servoing pipeline.

\begin{figure}
  \centering
  \includegraphics[width=0.48\textwidth]{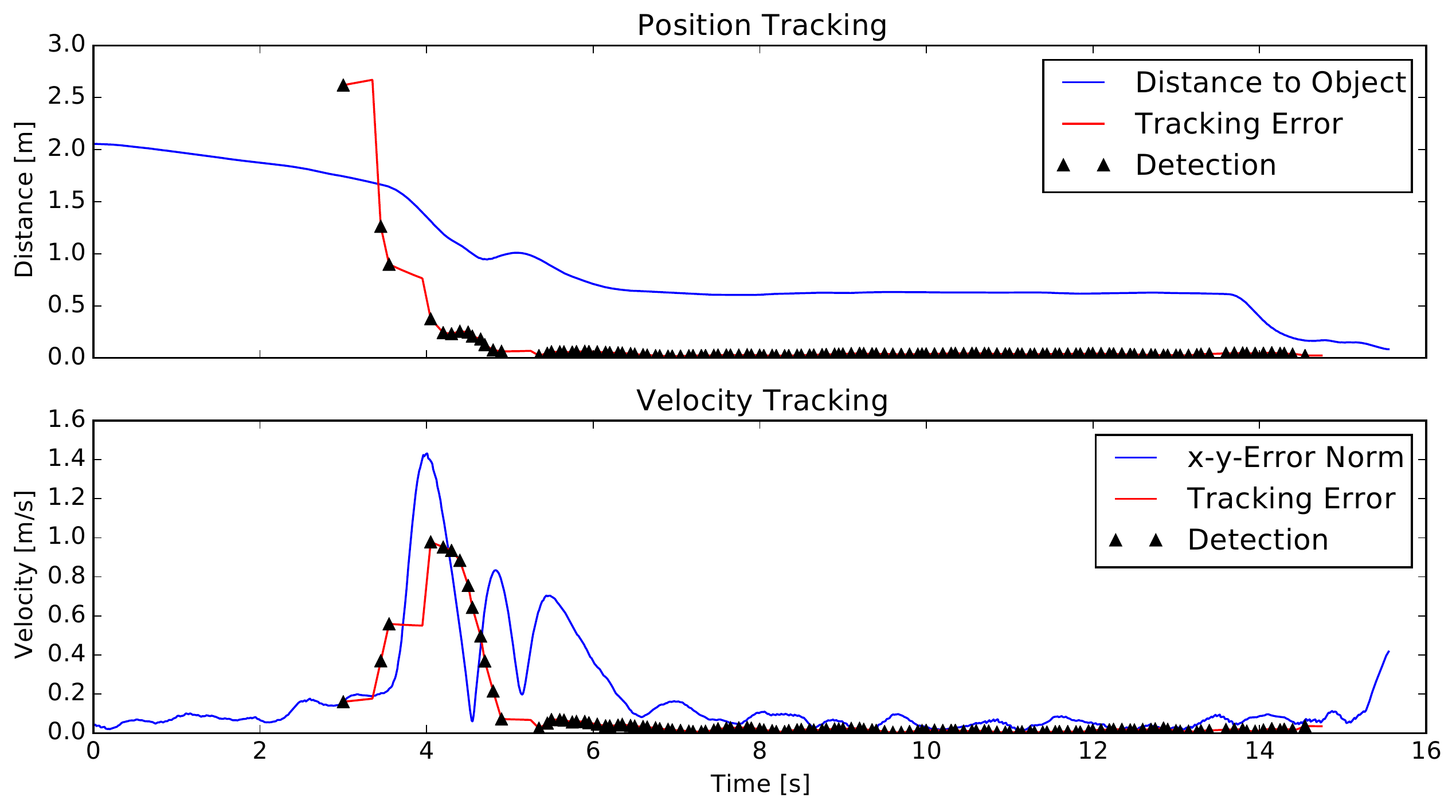}
  \caption{Tracking and picking up a moving object at $1 \unit{km/h}$.
  Once the object appears in the field of view, the \ac{mav} centers in a ball above it.
  The centering increases the detection rate and tracking accuracy.
  Detection errors on the border of the image cause initial tracking errors but the target is not lost.}
  \label{fig:servoing_result}
\end{figure}

\Figure{fig:servoing_result} shows the result of a single moving object pick-up.
The object moves at a constant velocity of $1 \unit{km/h}$ and enters the \ac{fov}.
Detection errors on the image corners (see \Figure{fig:detection_error}) cause tracking errors which converge against the ground-truth value when getting closer and centering above the object.
The \ac{mav} approaches and grabs the object after activating the magnet.

\subsection{\ac{mbzirc}}
We scored second place in the \ac{mbzirc} where we engaged up to three \acp{mav} simultaneously.
We were able to collect both moving and static objects and at least two objects in each trial.
On-site we only had two test slots and in total four trials to adjust our system to the environment.
Thus tools for creating trajectories and tuning detection thresholds were mandatory.
Also having a system as simple as possible was a great advantage in Abu Dhabi.

During the trials especially the servoing and control pipeline worked well.
If an agent detected an object it generally servoed on spot even in windy runs.

Unfortunately, the \ac{epm} turned out to be too weak to always connect through the paint layer of the objects and the contact sensing sometimes failed due to wiring or limited response times.
Furthermore, we faced CPU overload due to serialization of high resolution camera images.
Not only did this prevent us from logging any data for debugging,
but sometimes it even lead to delay and then divergence in the state estimation which made it hard to engage all \acp{mav} simultaneously.
The missing logs further prevented us from debugging detection issues which occured occasionally.
While we could not foresee the magnet force issues, more elaborate testing before the event may have revealed all other issues.

\section{Conclusion}
\label{sec:conclusions}
% Contribution
We presented a fully autonomous aerial system to search, pick up, and relocate objects.
Executing such complex task autonomously reduces operator effort and will eventually extend \acp{mav} employment in real-life scenarios such as \ac{sar}.
% Methods
Reactive collision avoidance makes our system decentralized such that each agent can act independently with minimum communication.
The agents are able to detect, track, and pick-up moving and static objects using monocular camera images and an \ac{epm} gripper.
Modularity allows deploying the system in different environments and performing only subtasks.
% Validation
We employed the system in various outdoor and indoor events and evaluated its components individually in ground-truth experiments.
% Results
Our system outperformed fifteen other systems at the \ac{mbzirc} where we became second.

\section*{Acknowledgment}
This work was supported by the Mohamed Bin Zayed International Robotics Challenge 2017,
the European Community's Seventh Framework Programme (FP7) under grant agreement n.608849 (EuRoC),
and the European Union's Horizon 2020 research and innovation programme under grant agreement n.644128 (Aeroworks).
\bibliographystyle{IEEEtran}
\bibliography{bibliography.bib}
\end{document}